\documentclass[11pt]{article}
\usepackage[final]{acl}

\usepackage{times}
\usepackage{latexsym}
\usepackage{booktabs}
\usepackage{tabularx}
\usepackage[normalem]{ulem}
\usepackage{tcolorbox}
\usepackage[T1]{fontenc}
\usepackage[utf8]{inputenc}
\usepackage{microtype}

\usepackage{inconsolata}
\usepackage{enumitem}
\usepackage{graphicx}
\usepackage{capt-of}

\title{The Impact of Editorial Intervention on Detecting Native Language Traces}

\author{
  \textbf{Ahmet Yavuz Uluslu}$^{1,2}$ \quad
  \textbf{Mark Gales}$^1$ \quad
  \textbf{Kate Knill}$^1$ \quad
  \textbf{Gerold Schneider}$^2$ \\[0.2cm]
  $^1$ALTA Institute, Department of Engineering, University of Cambridge \\
  $^2$Department of Computational Linguistics, University of Zurich \\[0.2cm]
  $^1$\texttt{\{mjfg, kmk1001\}@cam.ac.uk} \\
  $^2$\texttt{\{ahmetyavuz.uluslu, gerold.schneider\}@uzh.ch}
}

\begin{document}
\maketitle
\begin{abstract}
Native Language Identification (NLI) is the task of determining an author's native language (L1) from their non-native writing. With the advent of human-AI co-authorship, learner texts are routinely corrected and rewritten by large language models, fundamentally altering the linguistic features NLI approaches depend on. In this paper, we investigate the robustness of L1 traces across increasing degrees of editorial intervention. By processing 450 essays from the Write \& Improve 2024 (W\&I) corpus through varying levels of grammatical error correction and paraphrasing, we demonstrate that L1 attribution does not depend solely on surface-level errors. Instead, the detection models appear to leverage deeper L1-related features, including unidiomatic lexico-semantic choices and pragmatic transfer. We find that minimal edits preserve these structural traces and maintain high L1 attribution accuracy. In contrast, fluency edits and paraphrasing normalize these L1 features, leading to a sharp decline in performance.
\end{abstract}

\section{Introduction}
Native language identification (NLI) is the task of automatically identifying the native language (L1) of an individual based on a writing sample in a non-native language (L2). Cross-linguistic influence (CLI) is increasingly understood to be a persistent and likely inevitable characteristic of bilingual cognition \citep{uluslu2026neurocomputational}. Even highly proficient speakers retain unconscious traces of their L1 that are detectable across linguistic levels, from phonology to semantics to morphosyntax \citep{markov2022exploiting}. Applications of NLI span several fields, including forensic linguistics---for example, authorship attribution in cybercrime \citep{perkins2021application} and plagiarism detection \citep{wastl2025machine}---as well as second-language acquisition research focused on learner characteristics \citep{berti2023unravelling}. While early NLI systems relied heavily on hand-crafted linguistic features and machine learning algorithms \citep{malmasi2017report}, the recent transition toward large language models (LLMs) has established a new state-of-the-art (SoTA) \citep{ng2025leveraging}. This leap in zero-shot performance is often attributed to emergent capabilities from multilingual pretraining and related tasks, such as grammatical error detection \citep{zhang2023native}.

Widespread human-LLM co-authorship is forcing a fundamental shift in the data available for NLI \citep{huang2025authorship}. As non-native writers use generative AI tools to refine their work, learner errors that traditionally served as discriminative features for text analysis are becoming obsolete \citep{chen2025sok}. This evolution raises an important question: as LLM assistance moves from error correction to comprehensive edits, can NLI systems still detect an author’s native language? We investigate the robustness of these L1 traces across four levels of editorial intervention, ranging from unmodified learner writing to minimal GEC, unconstrained GEC incorporating fluency edits, and paraphrasing. Our findings reveal that L1 identification does not rely exclusively on grammatical errors; instead, NLI models can capture deeper linguistic phenomena, such as unidiomatic lexico-semantic choices and pragmatic transfer. While minimal-edit GEC leaves these underlying structures largely intact, paraphrasing acts as a normalizing filter that obscures L1-specific markers. Our results demonstrate that to remain viable in an era of widespread AI assistance, NLI research must develop methods to assess the degree of editorial intervention.

The rest of the paper is structured as follows. Section 2 reviews related work in NLI and the use of generative AI in writing. Section 3 describes our dataset and LLM editorial interventions. Section 4 details the experimental setup, Section 5 presents the quantitative results and qualitative feature analysis, and Section 6 concludes the paper.

\section{Related Work}
A recent survey of the NLI landscape indicates that many contemporary approaches have shifted toward adopting LLMs \citep{goswami2024native}. 
This aligns with broader trends in computational discourse analysis, where these models increasingly automate the inference of sociodemographic variables across digital platforms \citep{alizadeh2025web}. 
Subsequent research has investigated the performance and fine-tuning potential of various open-source models on NLI \citep{ng2025leveraging}. Other investigations have introduced masking strategies to isolate specific structural elements \citep{uluslu-schneider-2025-investigating} and examined the consistency of model explanations \citep{uluslu2025robust}. Additionally, studies have explored the capacity of these models to act as sophisticated feature extractors for detecting CLI—for instance, by tracing L1-induced word order or orthographic errors in L2 English directly back to L1 Arabic structures \citep{acharya-etal-2025-tracing}.

While LLMs have advanced the analytical capabilities of NLI systems, their concurrent rise as writing assistants challenges the assumptions underlying the task. Non-native writers are increasingly adopting generative AI tools across a spectrum of editorial intervention, ranging from targeted grammar checks to paraphrasing \citep{yang2024chatgpt}. From an NLI perspective, this shift can effectively neutralize the orthographic and morphosyntactic errors that historically served as the most discriminative features for L1 identification \citep{malmasi2017report}. To formalize this intervention, modern GEC is typically defined by two primary annotation paradigms \citep{bryant2023grammatical}. Minimal edits are strictly constrained to enforcing basic grammaticality while maintaining faithfulness to the author's original meaning and syntax. Fluency edits extend beyond targeted corrections, introducing broader structural changes to improve overall sentence flow and native-like phrasing. As texts advance along this continuum toward paraphrasing, the author's lexical and syntactic fingerprints are systematically overwritten \citep{richburg2024automatic}. The exact trajectory of NLI performance across this range of L1 feature retention remains underexplored. This paper addresses that gap by quantifying the impact of textual modification on NLI accuracy across editorial interventions, from unmodified learner texts to minimal edits, fluency edits, and paraphrasing.

\section{Data}
\subsection{W\&I 2024 NLI Dataset}
We evaluate our approach using a curated subset of the Write \& Improve (W\&I) 2024 corpus \citep{nicholls2024write}, which features manual minimal-edit GEC and CEFR-level annotations provided by human experts. To ensure our results remain directly comparable with existing literature, we construct our subset to mirror the linguistic diversity of the TOEFL11 dataset \citep{blanchard2013toefl11}, the de facto standard benchmark for NLI. Specifically, we select essays from nine L1 backgrounds available in W\&I that intersect with the TOEFL11 languages: Arabic, Chinese, French, German, Hindi, Italian, Japanese, Spanish, and Turkish. We sample 50 examples from each L1, yielding a balanced evaluation corpus of 450 texts. We restrict our selection to essays containing at least 100 words, aligning with established thresholds in stylometric and authorship studies \citep{koppel2011authorship}. Furthermore, we only include essays assessed at CEFR level A2 or above, ensuring the texts exhibit sufficient structural coherence and vocabulary for L1 feature analysis \citep{knill2025introducing}. After filtering, the final evaluation corpus has a mean essay length of $193 \pm 61$ words.

\begin{figure*}[t!]
    \centering
    \includegraphics[width=\textwidth]{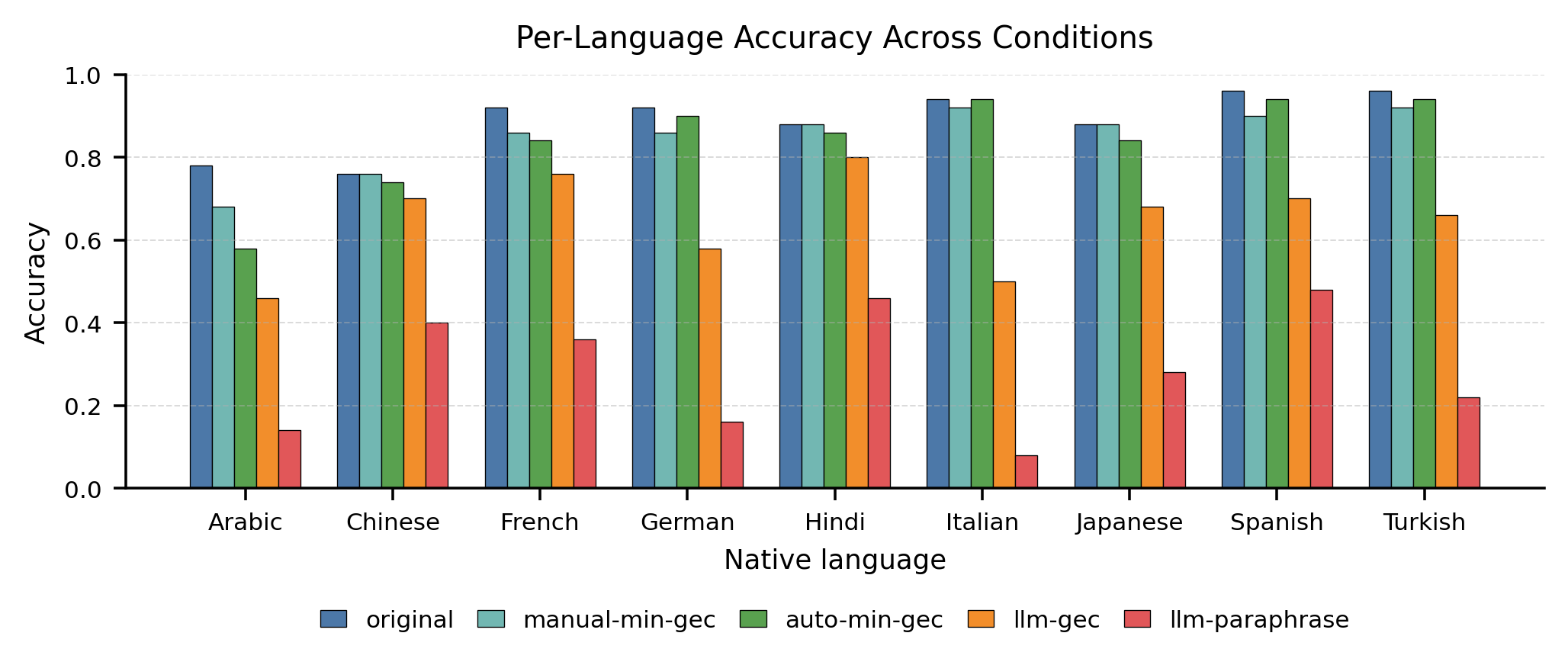}
    \caption{NLI accuracy scores across nine L1 backgrounds under varying degrees of editorial intervention.}
    \label{fig:class_accuracy_pagewide}
\end{figure*}


\begin{table}[ht!]
\centering
\small
\begin{tabularx}{\columnwidth}{@{}l >{\raggedright\arraybackslash}X@{}}
\toprule
\textbf{Stage} & \textbf{Sentence} \\
\midrule
original &
\textsuperscript{*}Every English learner searches the way how to learn
english while now a days most of learners searches online method. \\
\addlinespace
manual-min-gec &
Every English learner searches for the way to learn English while
nowadays most learners search for an online method. \\
\addlinespace
llm-gec &
Every English learner searches for ways to learn English, and nowadays,
most learners look for online methods. \\
\addlinespace
llm-paraphrase &
Every English learner seeks ways to improve their skills, and nowadays,
most turn to online methods. \\
\bottomrule
\end{tabularx}
\caption{Transformation of a W\&I sentence from a Hindi L1 writer across
editorial conditions. The auto-min-gec output is identical to the manual-min-gec output and is therefore omitted.}
\label{tab:wandi_searches_sentence}
\end{table}

\subsection{Editorial Interventions}
To investigate how varying degrees of textual modification affect NLI performance, we evaluate four levels of editorial intervention, instantiated by five experimental conditions. The minimal-edit level includes two parallel conditions---human expert correction and automatic minimal correction---allowing us to compare the two implementations directly. For our LLM-based interventions, we employ \texttt{gpt-4o-mini} as a representative baseline for a typical, widely accessible daily assistant capable of both GEC \citep{kovalchuk-etal-2025-introducing} and paraphrasing \citep{wang2025unveiling}. The exact prompts used for each condition are provided in Appendix~\ref{sec:prompts}. The five experimental conditions are defined as follows (see Table~\ref{tab:wandi_searches_sentence} for an illustrative example):

\begin{itemize}
    \item \textbf{original:} The unmodified learner texts extracted directly from the W\&I corpus, serving as our baseline for maximum L1 feature retention.
    \item \textbf{manual-min-gec:} Expert-annotated minimal corrections provided with the W\&I corpus \citep{nicholls2024write}.
    \item \textbf{auto-min-gec:} The learner texts processed by a SoTA minimal-edit system \citep{staruch-etal-2025-adapting} to determine if automated correction achieves parity with human expert annotations.

    \item \textbf{llm-gec:} A simple GEC setup in which the AI assistant is prompted to correct the grammatical errors in the text without edit restrictions. LLMs in this setting routinely over-correct, blending necessary fixes with subjective fluency edits \citep{zhan2026jelv}.
    
    \item \textbf{llm-paraphrase:} An unconstrained setup where the AI assistant is instructed to rewrite the text while preserving the original meaning and tone. This simulates real-world usage in which authors use AI assistants to refine their writing without completely abandoning its original style \citep{wang2025catch}.
\end{itemize}

\section{Methodology}
\subsection{L1 Classification}
We utilize OpenAI's \texttt{gpt-4o} for the NLI task, as it represents current SoTA performance \citep{goswami2025multilingual} and provides access to token-level probabilities. We additionally evaluate our approach using Qwen3.6-Plus \citep{qwen36plus} to assess whether the observed pattern generalizes to another model. We employ a zero-shot prompting paradigm, with the exact prompt templates provided in Appendix \ref{sec:prompts}. By formulating the task as a multiple-choice selection that maps nine candidate languages plus an `Other' category to letter identifiers, we apply constrained decoding to extract log probabilities and investigate model confidence (detailed in Appendix \ref{sec:uncertainty}). To mitigate position bias and obtain a more robust estimate of predictive uncertainty, we evaluate each instance across three randomized label orderings \citep{liusie2024llm}, applying a softmax over the logits of the valid candidate tokens and averaging the resulting distributions; the class with the highest aggregate probability is selected as the final prediction. Furthermore, to mitigate topic bias and prevent reliance on content-based shortcuts, we mask named entities (including locations, nationalities, and personal names) and non-English words prior to classification \citep{uluslu2025robust}.

\subsection{Evaluation Metrics}
We evaluate performance on the NLI task using standard classification accuracy. To characterize the extent and nature of textual modification across conditions, we compare each output with the human minimal-edit reference. Specifically, we report the MaxMatch ($M^2$) $F_{0.5}$ score \citep{dahlmeier2012better}, computed using the ERRANT toolkit \citep{bryant2017automatic}, together with Word Error Rate (WER). Additionally, we utilize BERTScore \citep{zhang2019bertscore} with a \texttt{microsoft/deberta-xlarge-mnli} backbone to assess semantic faithfulness. We use this metric to calculate the similarity between the manually corrected texts and the LLM-generated outputs (\texttt{llm-gec} and \texttt{llm-paraphrase}) and thereby assess the extent to which the original meaning is preserved.

\section{Results}

\subsection{Quantitative Performance}
Our experimental results, detailed in Table \ref{tab:m2_wer_gleu_main}, show that NLI performance declines as the degree of editorial intervention increases. The model (\texttt{gpt-4o}) achieves an initial accuracy of 88.9\% on the unmodified baseline texts. Applying minimal edits results in only a modest performance decline: accuracy decreases to 85.1\% for human corrections and 84.2\% for automatic corrections. This indicates that overt grammatical and orthographic errors are not the only cues supporting L1 identification.

\begin{table}[h!]
\centering
\begin{tabular}{lccc}
\hline
Condition & $M^{2}$ & WER\% & Accuracy\%\\
\hline
original                & ---    & 13.0 & 88.9 \\
auto-min-gec            & 0.54 & 9.7 & 84.2 \\
llm-gec                 & 0.25 & 26.3 & 64.9 \\
llm-paraphrase          & 0.11 & 54.2 & 28.7 \\
\hline
\end{tabular}
\caption{Comparison of text outputs with the human minimal-edit reference. $M^{2}$ reports F$_{0.5}$. Lower Word Error Rate (WER) indicates fewer edits relative to the reference. The \texttt{manual-min-gec} condition is omitted because it serves as the reference.}
\label{tab:m2_wer_gleu_main}
\end{table}

In the \texttt{llm-gec} condition, where the model applies subjective fluency edits alongside grammatical corrections, accuracy drops to 64.9\%. In the unconstrained \texttt{llm-paraphrase} setting, accuracy decreases to 28.7\%. Our supplementary evaluation using Qwen3.6-Plus demonstrates the same downward trend, with classification accuracy falling from 75.6\% on the original learner texts to 25.3\% under the paraphrasing condition. The results for this model are detailed in Table \ref{tab:models_classwise_calibrated_accuracy}. The decline in the performance of our primary model is consistent with the text-similarity metrics in Table~\ref{tab:m2_wer_gleu_main} and the model-confidence scores in Table~\ref{tab:main_results_uncertainty}. To further investigate the performance degradation in \texttt{llm-gec} and \texttt{llm-paraphrase} conditions, we provide the corresponding confusion matrices in Figure~\ref{fig:confusion_paraphrase}. As the Word Error Rate (WER) increases and the $M^2$ score decreases, indicating heavier deviation from the reference text, NLI accuracy falls correspondingly. Importantly, our similarity analysis suggests that the outputs retain strong semantic similarity to the manually corrected references; BERTScore values remain high in both the \texttt{llm-gec} ($F_1 = 0.924$) and \texttt{llm-paraphrase} ($F_1 = 0.847$) conditions.

\subsection{Qualitative Feature Analysis}
As illustrated in Table~\ref{tab:wandi_searches_sentence}, the original (unmodified) sentence from an L1 Hindi speaker contains several potential manifestations of cross-linguistic influence. At the surface level, it exhibits morphosyntactic errors, such as missing prepositions and articles (e.g., ``searches online method'') and subject--verb agreement problems (``most of learners searches''), which have been observed among English learners with an L1 Hindi background \citep{usama2024comparative}. The minimal-edit systems (\texttt{manual-min-gec} and \texttt{auto-min-gec}) correct these grammatical errors while preserving much of the learner's syntactic structure. The non-native-like phrasing \texttt{[the way] (+) [how to learn english]} may reflect structural transfer from Hindi correlative constructions \citep{liptak2009landscape}. The \texttt{llm-gec} output performs a broader revision, replacing this structure with ``searches for ways to learn English.''

To enable a closer examination of the features affected by editorial intervention, Appendix~\ref{sec:model-outputs} presents selected sentences from the same essay. The author repeatedly employs the discourse marker ``As we know.'' This framing presents the following proposition as shared knowledge; its repeated use may reflect pedagogical templates found in some South Asian educational contexts and is compatible with discourse patterns described for Indian English
\citep{valentine1991getting}. Similarly, phrases such as ``to live well in this world'' preserve literal or conceptually extended formulations that may derive from underlying L1 expressions. These features remain visible in the \texttt{original}, \texttt{auto-min-gec}, and \texttt{llm-gec} versions, and the model predicts Hindi for all three versions.

The \texttt{llm-paraphrase} condition, however, applies more extensive normalization. It reduces the repeated use of ``As we know,'' replaces literal formulations with more idiomatic alternatives, and introduces new lexical choices such as ``improve their skills'' and ``online methods.'' This process does not simply correct grammatical errors; it also removes the discourse framing and structural features that contribute to the text's profile. As a result, the classifier no longer predicts Hindi, instead predicting Chinese with low confidence.

\subsection{Error Analysis}
As the texts lose cues associated with the nine target classes, the model might be expected to assign an increasing proportion of texts to the `Other' category. However, the confusion matrices (Appendix \ref{sec:confusion_matrices}) indicate that this shift happens gradually. Under the intermediate \texttt{llm-gec} condition, the model primarily re-ranks the existing L1 classes. It is only under the \texttt{llm-paraphrase} condition that we observe a marked increase in `Other' predictions across all backgrounds. While this anticipated convergence is supported by the overall decrease in model confidence (Appendix \ref{sec:uncertainty}), the texts do not uniformly shift to this classification. Instead, \texttt{gpt-4o} exhibits a fallback behavior: when faced with high predictive uncertainty, the model frequently misclassifies the paraphrases as Spanish. Interestingly, a parallel evaluation using \texttt{Qwen3.6-Plus} demonstrates a similar fallback mechanism, but with Chinese as the most frequent prediction. We hypothesize that this phenomenon is driven by the models' pretraining corpora, which implicitly associate generic learner English with their most prevalent L2 demographics. This behavior may help explain why accuracy for Spanish remains relatively high at 48\% even under the \texttt{llm-paraphrase} condition. The residual accuracy for Chinese (40\%) and Hindi (46\%) may have a different explanation. As discussed in the previous section, LLM-based rewriting does not uniformly alter every linguistic feature. In some outputs, syntactic structures, lexical choices, and discourse patterns remain intact. These artifacts likely extend beyond syntax to include cultural and pragmatic elements that escape entity redaction. Anecdotal passages detailing a writer's perspective or cultural practices are difficult to entirely neutralize. 

\newpage

\section{Conclusion}
In this paper, we investigated the resilience of L1-related features under increasing levels of editorial intervention. By evaluating NLI performance across a continuum of textual modifications, we demonstrate that L1 attribution with LLMs does not rely solely on grammatical errors. Our findings reveal that while GEC remediates morphosyntactic violations, it often leaves lexico-semantic and conceptual traces, enabling high attribution accuracy. However, as the degree of intervention increases toward fluency edits and paraphrasing, these persistent features are increasingly normalized. This process effectively overwrites the author's linguistic profile, reducing NLI accuracy from 88.9\% to 28.7\%. As a result, future NLI research must establish methods to assess editorial intervention and prioritize the detection of persistent patterns---such as conceptual and pragmatic transfer---that survive the automated editing process.

\newpage

\section*{Limitations}
\label{sec:limitations}

\paragraph{One-Pass Intervention vs.\ True Co-Authorship}
Our experiments evaluate AI assistance as a single-pass, essay-level editorial intervention. This differs from authentic human-LLM co-authorship, which is highly iterative. In real-world scenarios, users often engage in a back-and-forth dialogue with the AI---rejecting specific edits, prompting for localized adjustments, or manually injecting their own phrasing and structural choices back into the revised text. 

\paragraph{Restriction to L2 English}
While our dataset covers a diverse set of nine L1 backgrounds, our evaluation is limited to English as the target language. The capabilities of both GEC and NLI systems are often weaker for target languages other than English \citep{goswami2025multilingual}. The dynamics of L1 feature erasure and the resilience of structural traces may manifest quite differently in non-English L2 contexts.

\section*{Ethics Statement}
Our research exclusively utilized a publicly available L2 English learner corpus: the pseudonymized W\&I corpus \citep{nicholls2024write}, which contains no personally identifiable information. We acknowledge the broad societal implications of authorship analysis, including potential risks to the security and privacy of individuals \cite{saxena2025responsible}. Our results are presented in a controlled experimental setting, primarily to study the impact of editorial intervention on existing NLI systems and explore approaches for enhancing their robustness. This work is not intended for deployment in critical real-world applications, such as forensic linguistics or automated academic integrity enforcement, without extensive further safeguards. As detailed in our \nameref{sec:limitations} section, we also recognize that our efforts to mitigate bias are not exhaustive, and further research is needed.

\section*{Acknowledgments}
This work was supported by the collaboration between the University of Zurich and PRODAFT as part of the Innosuisse innovation project 103.188 IP-ICT conducted at Linguistic Research Infrastructure (LiRI). The project also received support from Cambridge University Press \& Assessment, a department of The Chancellor, Masters, and Scholars of the University of Cambridge.

\bibliography{custom.bib}

\clearpage
\appendix
\counterwithin{figure}{section}
\counterwithin{table}{section}

\section{Experimental Prompts}
\label{sec:prompts}
For the generative editorial interventions (\textbf{llm-paraphrase} and \textbf{llm-gec}), the model was initialized with the standard system prompt: ``You are a helpful assistant.'' The L1 classification task utilized a specialized forensic expert persona.

\subsection{LLM-Paraphrase Prompt}
For the \textbf{llm-paraphrase} condition, the model was provided with the learner text and the following instruction to induce rewriting:
\begin{quote}
``Rewrite the text to sound natural and fluent, preserving meaning and tone. Don't add new information. Return only the rewritten text. \texttt{\textbackslash n\textbackslash n\{text\}}''
\end{quote}

\subsection{LLM-GEC Prompt}
For the \textbf{llm-gec} condition, the model was instructed to act as an unconstrained grammar corrector using the following prompt:
\begin{quote}
``Improve my English grammar and fix my errors. Return only the corrected text, with no explanation \texttt{\textbackslash n\textbackslash n\{text\}}.''
\end{quote}

\subsection{L1 Classification Prompt}
For the L1 attribution task, the model was cast in an expert persona and provided with a multiple-choice selection of the nine candidate languages (represented by \textit{\{options\}}). The exact prompt template was:
\begin{quote}
``You are a forensic linguistics expert that reads essays from non-native speakers of English in order to identify their native language. Use clues such as word choice, syntactic patterns, and grammatical errors to decide. Analyze the text and choose the most likely native language from the following options:

\{options\}

(X) Other

Reply with only the single letter (A, B, C, etc.) representing your choice.''
\end{quote}

\newpage

\section{Model Confidence and Uncertainty}
\label{sec:uncertainty}

We characterize model confidence and uncertainty using three metrics: mean true-class probability, mean predicted-class confidence, and predictive entropy
\citep{ovadia2019can}. 

\begin{table}[htbp]
\small
\centering
\setlength{\tabcolsep}{6pt} 
\begin{tabular}{lcccc}
\toprule
\textbf{Condition} & \textbf{Acc\%} & \textbf{$P(y)$} & \textbf{$P(\hat{y})$} & \textbf{Entropy} \\
\midrule
original              & 88.9 & 86.0 & 92.4 & 0.22 \\
manual-min-gec        & 85.1 & 79.4 & 86.9 & 0.37 \\
auto-min-gec          & 84.2 & 80.5 & 88.3 & 0.33 \\
llm-gec               & 64.9 & 58.0 & 73.5 & 0.76 \\
llm-paraphrase        & 28.7 & 25.1 & 56.4 & 1.20 \\
\bottomrule
\end{tabular}
\caption{Model performance and predictive uncertainty across experimental conditions. Metrics include Accuracy (Acc.), Mean True Probability $P(y)$, Mean Confidence $P(\hat{y})$, and Entropy (Ent.). Accuracy and probability scores are presented as percentages (\%).}
\label{tab:main_results_uncertainty}
\end{table}

\subsection{Mean Probability and Confidence}
For our classification task, we average the predicted probabilities across option-order permutations. Let $x$ represent the input text, $y$ the correct target class, and $\hat{y}$ the model's predicted class. For a given instance $x$, we calculate the expected probability assigned to $y$ by averaging across its three permutations. We define the mean true probability, $P(y)$, as this expected probability averaged across the entire test set. Similarly, we define the mean confidence, $P(\hat{y})$, as the expected probability assigned to $\hat{y}$ given $x$, also averaged across the test set. These metrics allow us to determine how confident the model remains in its predictions and in the correct attribution after varying levels of GEC and paraphrasing.

\subsection{Predictive Entropy}
To capture the overall uncertainty across the entire predictive distribution, we calculate the Shannon entropy \citep{shannon1948mathematical} over the averaged output probabilities. For a set of possible classes $\mathcal{Y}$, using the expected probability $\hat{p}(y \mid x)$ derived from the permutations, the predictive entropy $H$ is defined as:
\begin{equation}
H(Y \mid x) = -\sum_{y \in \mathcal{Y}} \hat{p}(y \mid x) \log \hat{p}(y \mid x)
\end{equation}
We compute entropy for each instance and report the mean across the evaluation set. Higher entropy values signify a flatter expected probability distribution, indicating that the model is highly uncertain and struggles to attribute the text to a specific L1.

\onecolumn

\section{Confusion Matrices}
\label{sec:confusion_matrices}

\begin{center}
    \includegraphics[
        width=\textwidth,
        height=0.55\textheight,
        keepaspectratio
    ]{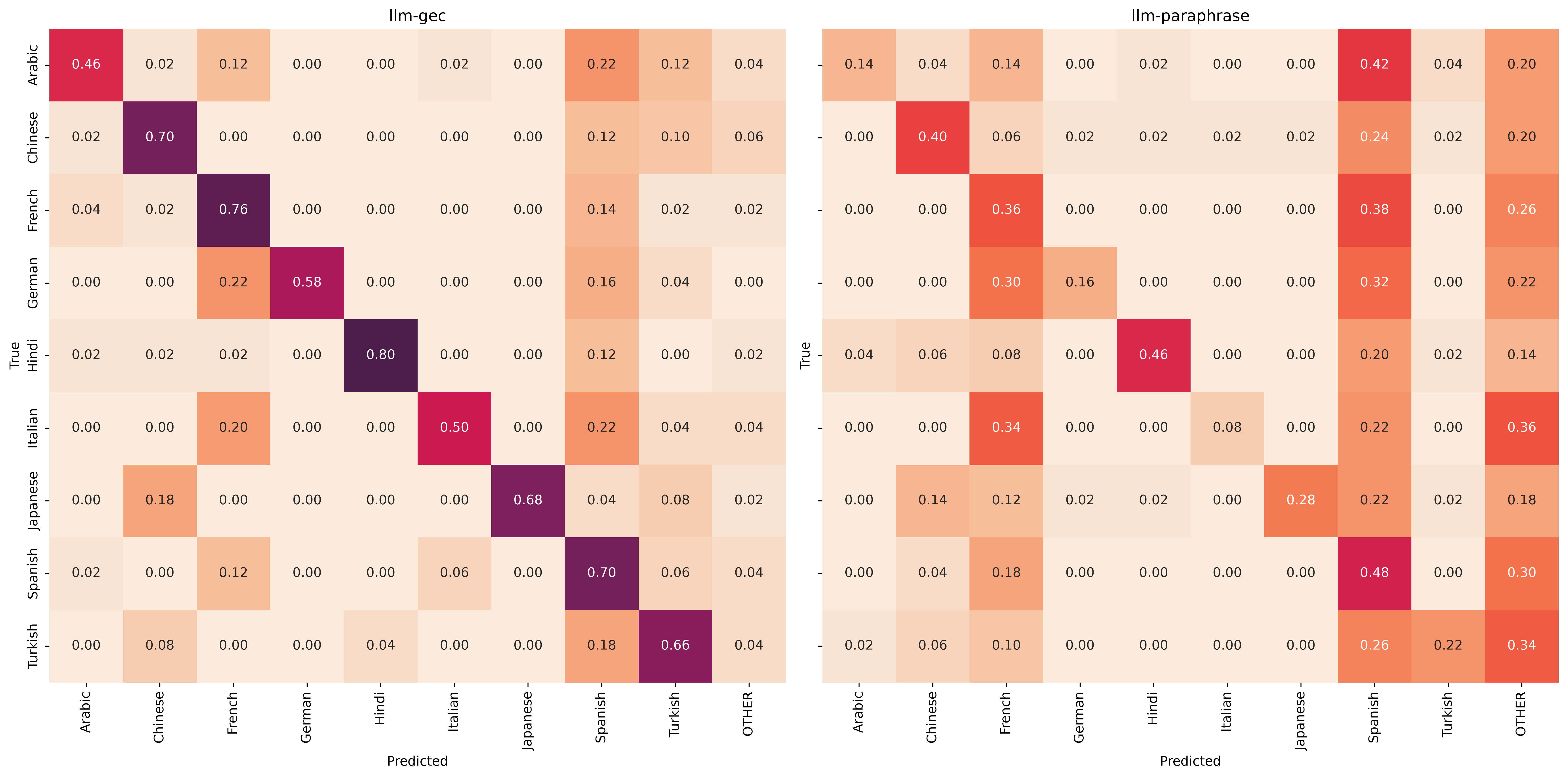}
    \captionof{figure}{Confusion matrices for GPT-4o L1 classification under the
    \texttt{llm-gec} and \texttt{llm-paraphrase} conditions.}
    \label{fig:confusion_paraphrase}
\end{center}

\vspace{0.5em}

\section{Supplementary Results}

\noindent
\begin{minipage}[t]{0.48\textwidth}
\centering
\small
\setlength{\tabcolsep}{8pt}
\begin{tabular}{lcc}
\toprule
 & \multicolumn{2}{c}{\textbf{Accuracy (\%)}} \\
\cmidrule(lr){2-3}
\textbf{Condition} & \textbf{Qwen3.6-Plus} & \textbf{GPT-4o} \\
\midrule
original         & 75.6 & 88.9 \\
manual-min-gec   & 65.6 & 85.1 \\
auto-min-gec     & 65.8 & 84.2 \\
llm-gec          & 44.0 & 64.9 \\
llm-paraphrase   & 25.3 & 28.7 \\
\bottomrule
\end{tabular}
\captionof{table}{Performance of Qwen3.6-Plus (calibrated) and GPT-4o across the five experimental conditions on the W\&I 2024 dataset.}
\label{tab:models_classwise_calibrated_accuracy}
\end{minipage}

\clearpage
\section{Model Outputs}
\label{sec:model-outputs}

\footnotesize
\setlength{\parindent}{0pt}
\setlength{\parskip}{0.45em}

\newtcolorbox{modelbox}[2]{
  colback=yellow!10!white,      
  colframe=gray!70!black,       
  colbacktitle=yellow!30!white, 
  coltitle=black,               
  fonttitle=\bfseries,          
  title={#1 \hfill \normalfont\small\textit{Classification: #2}}, 
  boxrule=0.5pt,                
  arc=3pt,                      
  left=4pt, right=4pt, top=4pt, bottom=4pt
}

\begin{figure}[htbp!]

\begin{modelbox}{original}{hindi}
\begin{enumerate}[label=(\arabic*), leftmargin=*, itemsep=0.45em, topsep=0pt]
    \item As we know English as secondary language is known as across word where it is not as first the language.
    \item As a language English is crucial to learn to live well in this world. A lot of issues which could be solved by knownig only English language.
    \item As we know, to learn any language, there are four parts, reading, writing, listening and speaking.
\end{enumerate}
\end{modelbox}

\vspace{0.5em}

\begin{modelbox}{auto-min-gec}{hindi}
\begin{enumerate}[label=(\arabic*), leftmargin=*, itemsep=0.45em, topsep=0pt]
    \item As we know, English as a secondary language is known as across word where it is not as the first language.
    \item As a language, English is crucial to learn to live well in this world. A lot of issues which could be solved by knowing only the English language.
    \item As we know, to learn any language, there are four parts: reading, writing, listening and speaking.
\end{enumerate}
\end{modelbox}

\vspace{0.5em}

\begin{modelbox}{llm-gec}{hindi}
\begin{enumerate}[label=(\arabic*), leftmargin=*, itemsep=0.45em, topsep=0pt]
    \item As we know, English as a secondary language is referred to as a foreign language when it is not the first language.
    \item As a language, English is crucial to learn in order to live well in this world. Many issues can be solved by knowing only the English language.
    \item As we know, to learn any language, there are four components: reading, writing, listening, and speaking.
\end{enumerate}
\end{modelbox}

\vspace{0.5em}

\begin{modelbox}{llm-paraphrase}{chinese}
\begin{enumerate}[label=(\arabic*), leftmargin=*, itemsep=0.45em, topsep=0pt]
    \item As we know, English is often referred to as a secondary language, meaning it is not the first language for many people.
    \item Learning English is essential for navigating the world effectively, as many issues can be resolved simply by knowing the language.
    \item To learn any language, we typically focus on four key areas: reading, writing, listening, and speaking.
\end{enumerate}
\end{modelbox}

\caption{Selected sentences authored by a Hindi L1 speaker from the W\&I 2024 corpus. The examples illustrate how various editorial interventions change the L1 classification (displayed in the top right of each panel).}

\end{figure}

\end{document}